\documentclass[letterpaper,10pt]{article}

\usepackage{epsfig}
\usepackage{url}

\usepackage{psfrag}
\usepackage{pst-all}
\usepackage{times}
\usepackage{epsfig}
\usepackage{graphicx}
\usepackage{amsmath}
\usepackage{amssymb}
\usepackage{algorithm}
\usepackage{algorithmic}
\usepackage{nameref}
\usepackage{subfigure}
\usepackage[numbers]{natbib}
\usepackage{fullpage}

\begin{document}

\title{A Supervised Neural Autoregressive Topic Model for Simultaneous Image Classification and Annotation}

\author{Yin Zheng\\
Department of Electronic Engineering, \\Tsinghua University, Beijing, China, 10084\\
{\tt\small y-zheng09@mails.tsinghua.edu.cn}
\and
Yu-Jin Zhang\\
Department of Electronic Engineering\\ Tsinghua University, Beijing, China, 10084\\
{\tt\small zhang-yj@mail.tsinghua.edu.cn}
\and
Hugo Larochelle\\
D\'{e}partment d'Informatique\\ Universit\'{e} de Sherbrooke, Sherbrooke (QC), Canada, J1K 2R1\\
{\tt\small hugo.larochelle@usherbrooke.ca }
}

\maketitle

\begin{abstract}
  Topic modeling based on latent Dirichlet allocation (LDA) has been a
  framework of choice to perform scene recognition and
  annotation. Recently, a new type of topic model called the Document
  Neural Autoregressive Distribution Estimator (DocNADE) was
  proposed and demonstrated state-of-the-art performance for document
  modeling. In this work, we show how to successfully apply and extend
  this model to the context of visual scene modeling.  Specifically,
  we propose SupDocNADE, a supervised extension of DocNADE, that
  increases the discriminative power of the hidden topic features by
  incorporating label information into the training objective of the
  model. We also describe how to leverage information about the spatial position
  of the visual words and how to embed additional image
  annotations, so as to simultaneously perform image classification
  and annotation. We test our model on the Scene15, LabelMe and UIUC-Sports
  datasets and show that it compares favorably to other
  topic models such as the supervised variant of LDA.

   
\end{abstract}

%
\section{Introduction}

Image classification and annotation are two important tasks in
computer vision. In image classification, one tries to describe the
image globally with a single descriptive label (such as {\it coast},
{\it outdoor}, {\it inside city}, etc.), while annotation focuses on tagging
the local content within the image (such as whether it contains ``{\it sky}'',
a ``{\it car}'', a ``{\it tree}'', etc.). Since these two problems are related, 
it is natural to attempt to solve them jointly. For example, an
image labeled as {\it street} is more likely to be annotated with ``{\it car}'',
``{\it pedestrian}'' or ``{\it building}'' than with ``{\it beach}'' or ``{\it see water}''. 
Although there has been a lot of work on image
classification and annotation separately, less work has looked at 
solving these two problems simultaneously.

Work on image classification and annotation is often based on a topic
model, the most popular being latent Dirichlet allocation or
LDA~\cite{blei2003latent}. LDA is a generative model for documents
that originates from the natural language processing community but
that has had great success in computer vision for scene
modeling~\cite{blei2003latent, wang2009simultaneous}. LDA models a
document as a multinomial distribution over topics, where a topic is
itself a multinomial distribution over words.  While the distribution
over topics is specific for each document, the topic-dependent
distributions over words are shared across all documents. Topic models
can thus extract a meaningful, semantic representation from a document
by inferring its latent distribution over topics from the words it
contains. In the context of computer vision, LDA can be used by first
extracting so-called ``visual words'' from images, convert the images
into visual word documents and training an LDA topic model on the
bags of visual words. Image representations learned with LDA have
been used successfully for many computer vision tasks such as visual
classification~\cite{lazebnik2006beyond,yang2009linear},
annotation~\cite{tsai2012bag,weston2010large} and image
retrieval~\cite{wu2009multi,philbin2007object}.

Although the original LDA topic model was proposed as an unsupervised
learning method, supervised variants of LDA have been
proposed~\cite{blei2007supervised, wang2009simultaneous}.  By modeling
both the documents' visual words and their class labels, the
discriminative power of the learned image representations could thus be
improved.

At the heart of most topic models is a generative story in which the
image's latent representation is generated first and the visual words
are subsequently produced from this representation. The appeal of this
approach is that the task of extracting the representation from
observations is easily framed as a probabilistic inference problem,
for which many general purpose solutions exist.  The disadvantage
however is that as a model becomes more sophisticated, inference
becomes less trivial and more computationally expensive. In LDA for
instance, inference of the distribution over topics does not have a
closed-form solution and must be approximated, either using
variational approximate inference or MCMC sampling. Yet, the model is
actually relatively simple, making certain simplifying independence
assumptions such the conditional independence of the visual words
given the image's latent distribution over topics.

Recently, an alternative generative modeling approach for documents
was proposed by \citet{larochelle2012neural}. Their model, the Document Neural
Autoregressive Distribution Estimator (DocNADE), models directly the
joint distribution of the words in a document, by decomposing it
through the probability chain rule as a product of conditional
distributions and modeling each conditional using a neural network. 
Hence, DocNADE doesn't incorporate any latent random
variables over which potentially expensive inference must be
performed. Instead, a document representation can be computed efficiently
in a simple feed-forward fashion, using the value of the neural
network's hidden layer. \citet{larochelle2012neural} also show that DocNADE is a better
generative model of text documents and can extract a useful 
representation for text information retrieval.

\begin{figure}[t]
\begin{center}
	\includegraphics[width=0.99\linewidth]{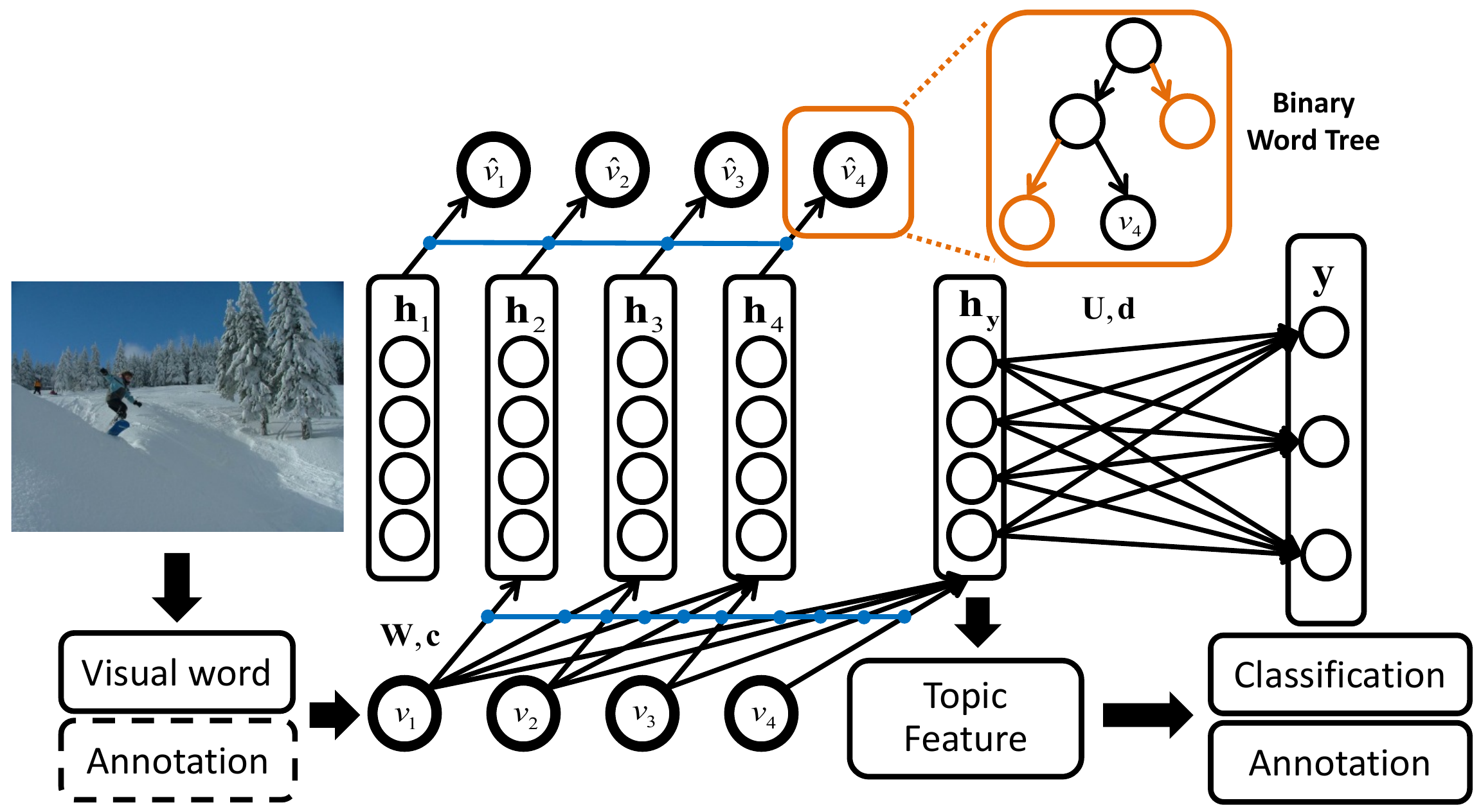}
\end{center}
\caption{Illustration of SupDocNADE for joint classification and
  annotation of images. Visual and annotation words are extracted
  from images and modeled by SupDocNADE, which models the joint
  distribution of the words ${\bf v} = [v_1,\dots,v_D]$ and class
  label $y$ as $p({\bf v},y) = p(y|{\bf v}) \prod_i
  p(v_i|v_1,\dots,v_{i-1})$.  All conditionals $p(y|{\bf v})$ and $
  p(v_i|v_1,\dots,v_{i-1})$ are modeled using neural networks with
  shared weights. Each predictive word conditional
  $p(v_i|v_1,\dots,v_{i-1})$ (noted ${\hat v}_i$ for brevity)
  follows a tree decomposition where each leaf is a possible word. At
  test time, the annotation words are not used (illustrated with a
  dotted box) to compute the image's topic feature representation.}
\label{fig:supdocnade}
\end{figure}

In this paper, we consider the application of DocNADE in the context
of computer vision. More specifically, we propose a supervised
variant of DocNADE (SupDocNADE), which models the joint distribution
over an image's visual words, annotation words and class label. The
model is illustrated in Figure~\ref{fig:supdocnade}. We investigate how to
successfully incorporate spatial information about the visual words
and highlight the importance of calibrating the generative and
discriminative components of the training objective. Our results
confirm that this approach can outperform the supervised variant
of LDA and is a competitive alternative for scene modeling.



\section{Related Work}
\label{relaged works}

Simultaneous image classification and annotation is often addressed
using models extending the basic LDA topic model.
\citet{wang2009simultaneous} proposed a supervised LDA formulation
to tackle this problem. \citet{wang2011max} opted instead for a
maximum margin formulation of LDA (MMLDA).  Our work also belongs to
this line of work, extending topic models to a supervised computer
vision problem: our contribution is to extend a different topic model,
DocNADE, to this context.

What distinguishes DocNADE from other topic models is its reliance on
a neural network architecture. Neural networks are increasingly used
for the probabilistic modeling of images (see~\cite{bengio2012representation} for
a review). In the realm of document modeling,
\citet{salakhutdinov2009replicated} proposed a Replicated Softmax model
for bags of words. DocNADE is in fact inspired by that model and was
shown to improve over its performance while being much more computationally
efficient. \citet{wanhybrid} also proposed a hybrid model that combines
LDA and a neural network. They applied their model to scene
classification only, outperforming approaches based on LDA or on a
neural network only. In our experiments, we show that our approach
outperforms theirs.  Generally speaking, we are not aware of any other
work which has considered the problem of jointly classifying and
annotating images using a hybrid topic model/neural network approach.


\section{Document NADE}
\label{DocNADE intro}


In this section, we describe the original DocNADE model. In \citet{larochelle2012neural},
DocNADE was use to model documents of real words, belonging to some
predefined vocabulary. To model image data, we assume that images have
first been converted into a bag of visual words. A
standard approach is to learn a vocabulary of visual words by
performing $K$-means clustering on SIFT descriptors densely
exacted from all training images. See Section~\ref{experiment:conf} for more details
about this procedure.  From that point on, any image can thus be
represented as a bag of visual words ${\bf v}=[v_1,v_2,\ldots,v_D]$, where each $v_i$ is the index of the
closest $K$-means cluster to the $i^{\rm th}$ SIFT descriptor
extracted from the image and $D$ is the number of extracted descriptors.

DocNADE models the joint probability of the visual words $p({\bf v})$ by rewritting it as
\begin{equation}
p\left({\bf v}\right)=\prod_{i=1}^{D} p\left ( v_i| {\bf v}_{<i}\right ) \label{eqn:prob_chain_rule}
\end{equation}
and modeling instead each conditional $p( v_i| {\bf v}_{<i} )$, where
$\mathbf{v}_{<i}$ is the subvector containing all $v_j$ such
that $j<i$.  Notice that Equation~\ref{eqn:prob_chain_rule} is true
for any distribution, based on the probability chain rule. Hence, the
main assumption made by DocNADE is in the form of the conditionals.
Specifically, DocNADE assumes that each conditional can be modeled and
learned by a feedforward neural network.

One possibility would be to model $ p( v_i| \mathbf{v}_{<i}) $ with the following architecture:
\begin{eqnarray}
\mathbf{h}_i\left ( \mathbf{v}_{<i} \right ) = {\bf g}\left( \mathbf{c}+\sum_{k<i}\mathbf{W}_{:,v_k} \right ) \label{eqn:docnade_hidden},&&
 p\left ( v_i=w|\mathbf{v}_{<i} \right ) = \frac{\exp\left ( b_w +\mathbf{V}_{w,:}\mathbf{h}_i\left ( \mathbf{v}_{<i} \right )\right )}{\sum_{w{}'}\exp\left ( b_{w{}'} +\mathbf{V}_{w{}',:}\mathbf{h}_i\left ( \mathbf{v}_{<i} \right )\right )}\label{eqn:docnade_softmax}
\end{eqnarray}
where $g(\cdot)$ is an element-wise non-linear activation function,
$\mathbf{W} \in \mathbb{R}^{H \times K}$ and $\mathbf{V} \in
\mathbb{R}^{K \times H}$ are the connection parameter matrices,
$\mathbf{c} \in \mathbb{R}^N$ and $\mathbf{b} \in \mathbb{R}^K$ are
bias parameter vectors and $H,K$ are the number of hidden units
(topics) and vocabulary size, respectively.

Computing the distribution $p( v_i=w|\mathbf{v}_{<i} )$ of
Equation~\ref{eqn:docnade_softmax} requires time linear in $K$. In
practice, this is too expensive, since it must be computed for each of
the $D$ visual words $v_i$.  To address this issue,
\citet{larochelle2012neural} propose to use a balanced binary tree to
decompose the computation of the conditionals and obtain a complexity
logarithmic in $K$. This is achieved by randomly assigning all visual
words to a different leaf in a binary tree. Given this tree, the
probability of a word is modeled as the probability of reaching its
associated leaf from the root. We model each left/right transition
probabilities in the binary tree using a set of binary logistic
regressors taking the hidden layer $\mathbf{h}_{i}({\bf v}_{<i})$ as
input. The probability of a given word can then be obtained by
multiplying the probabilities of each left/right choices of the
associated tree path.

Specifically, let $\mathbf{l}\left(v_i\right) $ be the sequence of tree
nodes on the path from the root to the leaf of $v_i$ and let $\pi
\left(v_i\right)$ be the sequence of binary left/right choices 
at the internal nodes along that path. For example, $l\left(v_i\right)_1 $
will always be the root node of the binary tree, and $\pi
\left(v_i\right)_1$ will be $0$ if the word leaf $v_i$ is in the left
subtree or $1$ otherwise. Let $\mathbf{V} \in \mathbb{R}^{T\times
  H} $ now be the matrix containing the logistic regression weights and
$\mathbf{b} \in \mathbb{R}^T$ be a vector containing the 
biases, where $T$ is the number of inner nodes in the binary tree
and $H$ is the number of hidden units. The probability $ p(
  v_i=w|\mathbf{v}_{<i} )$ is now modeled as
\begin{equation}
p( v_i=w|\mathbf{v}_{<i}) = \prod_{k=1}^{|\pi\left(v_i\right)|} p(\pi\left(v_i\right)_k|\mathbf{v}_{<i})~, \label{eqn:docnade_tree}
\end{equation}      
where
\begin{equation}
p(\pi\left(v_i\right)_k=1|\mathbf{v}_{<i})=\textup{ sigm}\left( b_{l\left(v_i\right)_m} +\mathbf{V}_{l\left(v_i\right)_m,:}\mathbf{h}_i\left ( \mathbf{v}_{<i} \right )\right)
\label{eqn:docnade_tree_lr}
\end{equation}
are the internal node logistic regression outputs and $\textup{
  sigm}(x) = 1/(1+\exp(-x))$ is the sigmoid function. By using a
balanced tree, we are guaranteed that computing
Equation~\ref{eqn:docnade_tree} involves only $O(\log K)$ logistic
regression outputs. One could attempt to optimize the organization of
the words within the tree, but a random assignment of the words to
leaves works well in practice \cite{larochelle2012neural}.

Thus, by combining Equations~\ref{eqn:docnade_hidden},
\ref{eqn:docnade_tree} and \ref{eqn:docnade_tree_lr}, we can compute
the probability $p\left( {\bf v} \right)=\prod_{i=1} p\left ( v_i|{\bf
    v}_{<i}\right ) $ for any document under DocNADE. To train the
parameters $\theta = \lbrace{{\bf W},{\bf V},{\bf b},{\bf c}\rbrace}$
of DocNADE, we simply optimize the average negative log-likelihood of
the training set documents using stochastic gradient descent. Once the
model is trained, a latent representation can be extracted from a new
document $\mathbf{v^{\ast}}$ as follows:
\begin{equation}
\mathbf{h}_y\left ( \mathbf{v}^{\ast} \right ) = {\bf g}\left( \mathbf{c}+\sum_{i}^{D}\mathbf{W}_{:,v^{\ast}_i} \right )~.
\end{equation} 
This representation could be fed to a standard classifier to perform
any supervised computer vision task. The index $y$ is used to
highlight that it is the representation used to predict the class
label $y$ of the image.

Equations~\ref{eqn:docnade_tree},\ref{eqn:docnade_tree_lr} indicate that the
conditional probability of each word $v_i$ requires computing the
position dependent hidden layer $\mathbf{h}_i\left( \mathbf{v}_{<i}
\right )$, which extracts a
representation out of the bag of previous visual words
$\mathbf{v}_{<i}$. Since computing $\mathbf{h}_i\left( \mathbf{v}_{<i}
\right )$ is in $O(H D)$ on average, and there are $D$
hidden layers $\mathbf{h}_i\left( \mathbf{v}_{<i}
\right )$ to compute, then a naive procedure
for computing all hidden layers would be in $O(H D^2)$.

However, noticing that
\begin{eqnarray}
{\mathbf{h}_{i+1}\left ( \mathbf{v}_{<i+1} \right )} & =& {\bf g}\left( \mathbf{c}+\sum_{k<i+1}\mathbf{W}_{:,v_k} \right )  =  {\bf g}\left( \mathbf{W}_{:,v_i}+\mathbf{c}+\sum_{k<i}\mathbf{W}_{:,v_k} \right )
\end{eqnarray}
and exploiting that fact that the weight matrix $\mathbf{W}$ is the same
across all conditionals, the linear transformation
$\mathbf{c}+\sum_{k<i}\mathbf{W}_{:,v_k}$ can be reused from the computation
of the previous hidden layer $\mathbf{h}_{i}( \mathbf{v}_{<i})$ to compute
$\mathbf{h}_{i+1}( \mathbf{v}_{<i+1})$. With this procedure,
computing all hidden layers $\mathbf{h}_{i}( \mathbf{v}_{<i}
) $ sequentially from $i=1$ to $i=D$ becomes in $O(H D)$.
 
Finally, since the computation complexity of each of the $O(\log K)$ logistic
regressions in Equation~\ref{eqn:docnade_tree} is $O(H)$, the total
complexity of computing $ p( v_i=w|\mathbf{v}_{<i} )$ is $O(\log(K) H
D)$. In practice, the length of document $D$ and the number of hidden
units $H$ tends to be small, while $\log(K)$ will be small even for large
vocabularies. Thus DocNADE can be used and trained efficiently.

\section{SupDocNADE for Image Classification and Annotation}
\label{SupDocNADE intro}
In this section, we describe the approach of this paper, inspired by DocNADE, to
simultaneously classify and annotate image data. First, we describe a
supervised extension of DocNADE (SupDocNADE), which incorporates class
label information into training to learn more discriminative hidden
features for classification. Then we describe how we exploit
the spatial position information of the visual words. At
last, we describe how to also perform annotation, along with classification,
using SupDocNADE.


\subsection{Supervised DocNADE}

It has been observed that learning image feature representations using
unsupervised topic models such as LDA can perform worse than training
a classifier directly on the visual words themselves, using an
appropriate kernel such as a pyramid
kernel~\cite{lazebnik2006beyond}. One reason is that the unsupervised
topic features are trained to explain as much of the entire
statistical structure of images as possible and might not model
well the particular discriminative structure we are after in our
computer vision task. This issue has been addressed in the literature
by devising supervised variants of LDA, such as Supervised
LDA or sLDA~\cite{blei2007supervised}. DocNADE also being an unsupervised
topic model, we propose here a supervised variant of DocNADE,
SupDocNADE, in an attempt to make the learned image representation
more discriminative for the purpose of image classification.

Specifically, given an image $ {\bf v}=[
v_1,v_2,\ldots,v_D]$ and its class label $y\in \{1,\dots,C\}$,
SupDocNADE models the full joint distribution as
\begin{equation}
p( {\bf v})=p(y|{\bf v})\prod_{i=1}^{D} p\left ( v_i| {\bf v}_{<i}\right ) ~~. \label{eqn:supdocnade}
\end{equation}
As in DocNADE, each conditional is modeled by a neural network. We use
the same architecture for $p\left ( v_i| {\bf v}_{<i}\right )$ as in regular DocNADE. We now only need
to define the model for $p(y|{\bf v})$.

Since $\mathbf{h}_y\left ( \mathbf{v} \right )$ is the image
representation that we'll use to perform classification, we propose
to model $p\left ( y| {\bf v}\right )$ as a multiclass logistic 
regression output computed from $\mathbf{h}_y\left ( \mathbf{v} \right )$:
\begin{equation}
p\left( y|{\bf v}\right) = {\rm softmax}\left( \mathbf{d} + \mathbf{U}\mathbf{h}_y\left (\mathbf{v} \right) \right)_y\label{eqn:h_class}
\end{equation}
where ${\rm softmax}({\bf a})_i = \exp(a_i) / \sum_{j=1}^C \exp(a_j)$,
 $ \mathbf{d} \in \mathbb{R}^C $ is the bias parameter vector in
the supervised layer and $ \mathbf{U} \in \mathbb{R}^{C \times H} $ is
the connection matrix between hidden layer $\mathbf{h}_y $ and the class label.

Put differently, $p\left ( y| {\bf v}\right )$ is modeled as a regular
multiclass neural network, taking as input the bag of visual words
${\bf v}$. The crucial difference however with a regular neural
network is that some of its parameters (namely the hidden unit
parameters ${\bf W}$ and ${\bf c}$) are also used to model the visual
word conditionals $p\left ( v_i| {\bf v}_{<i}\right )$.

Maximum likelihood training of this model is performed by
minimizing the negative log-likelihood
\begin{equation}
  -\log p\left( {\bf v},y\right) = - \log p\left ( y| {\bf v} \right) +\label{eqn:objectfunc}  \sum_{i=1}^{D} -\log p( v_i | {\bf v}_i)
\end{equation} 
averaged over all training images. This is known as generative
learning~\cite{bouchard2004tradeoff}. The first term is a purely discriminative term,
while the second is unsupervised and can be understood as a
regularizer, that encourages a solution which also explains the
unsupervised statistical structure within the visual words. In
practice, this regularizer can bias the solution too strongly away
from a more discriminative solution that generalizes well. Hence, 
similarly to previous work on hybrid generative/discriminative learning, we propose
instead to weight the importance of the generative term
\begin{equation}
-\log p\left( {\bf v},y\right) = - \log p\left ( y| {\bf v} \right) + \lambda \sum_{i=1}^{D} -\log p( v_i | {\bf v}_i) \label{eqn:objectfunc_hybrid} 
\end{equation} 
where $\lambda$ is treated as a regularization hyper-parameter.

Training on the training set average of
Equation~\ref{eqn:objectfunc_hybrid} is performed by stochastic
gradient descent, using backpropagation to compute the parameter
derivatives. As in regular DocNADE, computation of the training objective and its
gradient requires that we define an ordering of the visual
words. Though we could have defined an arbitrary path across the image
to order the words (e.g. from left to right, top to bottom in the
image), we follow~\citet{larochelle2012neural} and randomly permute the words before every
stochastic gradient update. The implication is that the model is
effectively trained to be a good inference model of {\it any} conditional $p(
v_i | {\bf v}_{<i})$, for any ordering of the words in ${\bf v}$. This
again helps fighting against overfitting and better regularizes our
model.

In our experiments, we used the rectified linear function as the
activation function
\begin{equation}
{\bf g}({\bf a}) = \max(0, {\bf a}) = [\max(0,a_1),\dots,\max(0,a_H)]
\end{equation}
which often outperforms other activation functions~\cite{glorot2011deep} and
has been shown to work well for image data~\cite{nair2010rectified}. Since
this is a piece-wise linear function, the (sub-)gradient
with respect to its input, needed by backpropagation to compute the parameter gradients, is simply
\begin{equation}
{\bf 1}_{({\bf g}({\bf a})>0)}= [1_{(g(a_1)>0)},\dots,1_{(g(a_H)>0)}]
\end{equation}
where $1_{P}$ is 1 if $P$ is true and 0 otherwise.
Algorithms~\ref{alg:fprop}~and~\ref{alg:bprop} give pseudocodes
for efficiently computing the joint distribution $p\left({\bf v},y \right)$
and the parameter gradients of Equation~\ref{eqn:objectfunc_hybrid} required
for stochastic gradient descent training.

\begin{algorithm}[t]
\caption{ Computing $p\left({\bf v},y \right)$ using SupDocNADE}
\begin{algorithmic}
\STATE {\bf Input:} bag of words representation ${\bf v}$, target $y$
\STATE {\bf Output:} $p\left({\bf v},y \right)$
\STATE $\mathbf{a}\gets \mathbf{c}$ 
\STATE $p\left(\mathbf{v} \right) \gets 1$ 
\FOR{$i$ from $1$ to $D$}
  \STATE $\mathbf{h}_i \gets$ ${\bf g}\left( \mathbf{a}\right)$
  \STATE $p\left(v_i|\mathbf{v}_{<i}\right)=1$ 
  \FOR{$m$ from 1 to $|\pi \left(v_i\right)|$}
    \STATE  $p\left(v_i|\mathbf{v}_{<i}\right) \gets  p\left(v_i|\mathbf{v}_{<i}\right) p\left(\pi\left(v_i\right)_m|\mathbf{v}_{<i}\right)$
  \ENDFOR
  \STATE $p\left(\mathbf{v} \right) \gets p\left(\mathbf{v} \right)p\left(v_i|\mathbf{v}_{<i}\right)$ 
  \STATE $\mathbf{a}\gets \mathbf{a} + \mathbf{W}_{:,v_i}$ 
\ENDFOR
\STATE $\mathbf{h}^{c}\left (\mathbf{v} \right ) \gets \max(0,\mathbf{a}) $
\STATE $p\left( y|\mathbf{v}\right) \gets \textup{softmax} \left( \mathbf{d} + \mathbf{U}\mathbf{h}^{c}\left (\mathbf{v} \right ) )\right)_{|y}$
\STATE $p\left({\bf v},y \right) \gets p\left(\mathbf{v} \right)p\left( y|\mathbf{v}\right) $
\end{algorithmic}
\label{alg:fprop}
\end{algorithm}

\begin{algorithm}[t]
\caption{ Computing SupDocNADE training gradients}
\begin{algorithmic}
\STATE {\bf Input:} training vector ${\bf v}$, target $y$,\\
\hspace{10mm} unsupervised learning weight $\lambda$
\STATE {\bf Output:} gradients of Equation~\ref{eqn:objectfunc_hybrid} w.r.t. parameters
\STATE $f\left(\mathbf{v}\right) \gets \textup{softmax} \left( \mathbf{d} + \mathbf{U}\mathbf{h}^{c}\left (\mathbf{v} \right ) )\right)$
\STATE $\delta \mathbf{d} \gets \left(f \left(\mathbf{v}\right)-1_y\right)$
\STATE $\delta \mathbf{a} \gets (\mathbf{U}^\intercal \delta \mathbf{d})  \circ 1_{{\bf h}_y > 0}$
\STATE $\delta \mathbf{U} \gets \delta \mathbf{d}~{\mathbf{h}^{c^\intercal}}$
\STATE $\delta \mathbf{c} \gets 0$, $\delta \mathbf{b} \gets 0$, $\delta \mathbf{V} \gets 0$, $\delta \mathbf{W} \gets 0$
\FOR{$i$ from $D$ to $1$}
   \STATE $\delta\mathbf{h}_i \gets 0$
   \FOR {$m$ from $1$ to $|\pi\left(v_i\right)|$}
   \STATE $ \delta t \gets \lambda \left(p\left(\pi\left(v_i\right)_m|\mathbf{v}_{<i}\right)-\pi\left(v_i\right)_m\right)$
   \STATE $\delta b_{l\left(v_i\right)_m} \gets \delta b_{l\left(v_i\right)_m}+\delta t$
   \STATE $\delta \mathbf{V}_{l\left(v_i\right)_m,:} \gets \delta \mathbf{V}_{l\left(v_i\right)_m,:}+ \delta t~\mathbf{h}_i^\intercal$
   \STATE $\delta \mathbf{h}_i \gets \delta \mathbf{h}_i + \delta t~\mathbf{V}_{l\left(v_i\right)_m,:}^\intercal$
   \ENDFOR
\STATE $\delta \mathbf{a} \gets \delta \mathbf{a}+ \delta \mathbf{h}_i\circ 1_{{\bf h}_i > 0}$
\STATE $\delta \mathbf{c} \gets \delta \mathbf{c} + \delta \mathbf{h}_i\circ 1_{{\bf h}_i > 0}$
\STATE $\delta \mathbf{W}_{:,v_i} \gets \delta \mathbf{W}_{:,v_i} + \delta \mathbf{a}$
\ENDFOR

\end{algorithmic}
\label{alg:bprop}
\end{algorithm}

\subsection{Dealing with Multiple Regions}
\label{sec:multiple regions}
Spatial information plays an important role for understanding an
image. For example, the sky will often appear on the top part
of the image, while a car will most often appear at the
bottom. A lot of previous work has exploited this intuition
successfully. For example, in the seminal work on spatial
pyramids~\cite{lazebnik2006beyond}, it is shown that extracting
different visual word histograms over distinct regions instead of a
single image-wide histogram can yield substantial gains in
performance.

We follow a similar approach, whereby we model both the presence of the
visual words and the identity of the region they appear in. 
Specifically, let's assume the image is divided into several distinct regions
${\cal R} = \lbrace{ R_1,R_2, \ldots, R_M \rbrace }$, where $M$ is the number of
regions. The image can now be represented as
\begin{eqnarray}
{\bf v}^{\cal R}& =& [ v^{\cal R}_1,v^{\cal R}_2, \ldots, v^{\cal R}_D ] = [ \left( v_1,r_1 \right),\left( v_2, r_2 \right),\ldots,\left(v_D,r_D \right) ]
\end{eqnarray} 
where $r_i \in {\cal R}$ is the region
from which the visual word $v_i$ was extracted. To model the joint
distribution over these visual words, we decompose it as $p({\bf v}^{\cal R}) =
\prod_i p((v_i,r_i) | {\bf v}^{\cal R}_{<i})$ and treat each $K\times M$
possible visual word/region pair as a distinct word. One implication
of this is that the binary tree of visual words must be larger so as
to have a leaf for each possible visual word/region pair. Fortunately,
since computations grow logarithmically with the size of the tree,
this is not a problem and we can still deal with a large
number of regions.

\subsection{Dealing with Annotations}

The annotation of an image consists in a list of
words\footnote{Annotations contain multiword expressions as well such
  as {\it person walking}, but they are treated as a single token.}
describing the content of the image. For example, in the image of
Figure~\ref{fig:supdocnade}, the annotation might contain the words
``{\it trees}'' or ``{\it people}''.  Because annotations and labels are clearly
dependent, we try to model them jointly within our SupDocNADE
model.

Specifically, let ${\cal A}$ be the predefined vocabulary of all
annotation words, we will note the annotation of a given image as $
{\bf a}= [ a_1,a_2, \ldots ,a_L ] $ where $a_i \in
{\cal A}$, with $L$ being the number of words in the annotation. Thus,
the image with its annotation can be represented as a mixed bag of visual
and annotation words:
\begin{eqnarray}
{\bf v}^{\cal A} & = & [ v_1^{\cal A},\ldots,v_D^{\cal A} , v_{D+1}^{\cal A}, \ldots, , v_{D+L}^{\cal A}] =  [ v^{\cal R}_1,\ldots,v^{\cal R}_D , a_1, \ldots ,a_L ]~~. 
\end{eqnarray}
To embed the annotation words into the SupDocNADE framework, we treat
each annotation word the same way we deal with visual
words. Specifically, we use a joint indexing of all visual and
annotation words and use a larger binary word tree so as to augment it with
leaves for the annotation words. By training SupDocNADE on this
joint image/annotation representation ${\bf v}^{\cal A}$, 
it can learn the relationship between the labels, the spatially-embedded 
visual words and the annotation words.

At test time, the annotation words are not given and we wish to
predict them. To achieve this, we compute the document representation
${\bf h}_y({\bf v}^{\cal R})$ based only on the visual words and
compute for each possible annotation word $a \in {\cal A}$ the
probability that it would be the next observed word $p(v_i^{\cal A} = a
|{\bf v}^{\cal A} = {\bf v}^{\cal R})$, based on the tree decomposition as in 
Equation~\ref{eqn:docnade_tree}. In other words, we only compute the probability
of paths that reach a leaf corresponding to an annotation word (not a visual word).
We then rank the annotation words in ${\cal A}$ in decreasing order of their probability
and select the top 5 words as our predicted annotation. 

\section{Experiments and Results}
\label{experiment}
In this section, we test our model on 3 real-world datasets: a subset
of the LabelMe dataset~\cite{russell2008labelme}, the UIUC-Sports dataset~\cite{li2007and} and the
Scene15 dataset~\cite{lazebnik2006beyond}. Scene15 is used to evaluate image
classification performance only, while LabelMe and UICU-Sports come
with annotations and is a popular classification and annotation
benchmark. We provide a quantitative comparison between SupDocNADE,
the original DocNADE model and supervised LDA
(sLDA)~\cite{blei2007supervised,wang2009simultaneous}.  The code to
download the datasets and for SupDocNADE is available at
\url{http://www.anonymous.com}.

\subsection{Datasets Description}

The Scene15 dataset contains 4485 images, belonging to 15
different classes. Following previous work, we first resize the images so 
the maximum side (length or width) is 300 pixels wide, without changing the aspect
ratio. For each experiment, we randomly select 100 images as the training set,
using the remaining images for the test set.
 
Following \citet{wang2009simultaneous}, we constructed our LabelMe
dataset using the online tool to obtain images of size $256 \times 256$ pixels from the
following 8 classes: {\it highway}, {\it inside city}, {\it coast}, {\it forest},
{\it tall building}, {\it street}, {\it open country} and {\it mountain}. For each
class, 200 images were randomly selected and split evenly in the training
and test sets, yielding a total of 1600 images. 

The UIUC-Sports dataset contains 1792 images, classified into
8 classes: {\it badminton} (313 images), {\it bocce} (137 images), {\it croquet}
(330 images), {\it polo} (183 images), {\it rockclimbing} (194 images), {\it rowing}
(255 images), {\it sailing} (190 images), {\it snowboarding} (190
images). Following previous work, the maximum side of each image was
resized to 400 pixels, while maintaining the aspect ratio. We
randomly split the images of each class evenly into training and test sets. 
For both LabelMe and UIUC-Sports
datasets, we removed the annotation words occurring less than 3 times, as
in \citet{wang2009simultaneous}.

\subsection{Experimental Setup}
\label{experiment:conf}

Following \citet{wang2009simultaneous}, 128
dimensional, densely extracted SIFT features were used to extract the
visual words. The step and patch size of the dense SIFT extraction was
set to 8 and 16, respectively. The dense SIFT features from the
training set were quantized into 240 clusters, to construct our visual
word vocabulary, using $K$-means.  We divided each image into
a $2 \times 2 $ grid to extract the spatial position information, as described in
Section~\ref{sec:multiple regions}.  This produced
$2\times 2\times 240=960$ different visual word/region pairs.

We use classification accuracy to evaluate the performance of
image classification and the average F\textup{-measure} of the top
5 predicted annotations to evaluate the annotation performance, as
in previous work. The F\textup{-measure} of an image is defined as
\begin{equation}
F\textup{-measure} = \frac{2\times \textup{Precision}\times \textup{Recall}}{ \textup{Precision}+\textup{Recall}}
\end{equation}
where recall is the percentage of correctly predicted annotations out
of all ground-truth annotations for an image, while the precision is
the percentage of correctly predicted annotations out of all predicted
annotations\footnote{When there are repeated words in the ground-truth
  annotations, the repeated terms were removed to calculate the
  F\textup{-measure}}. We used 5 random train/test splits to estimate
the average accuracy and F\textup{-measure}.

Image classification with SupDocNADE is performed by feeding the learned
document representations to a RBF kernel SVM. In our experiments, all
hyper-parameters (learning rate, unsupervised learning weight
$\lambda$ in SupDocNADE, $C$ and $\gamma$ in RBF kernel SVM), were
chosen by cross validation. We emphasize that the annotation words are
not available at test time and all methods predict an image's class
based solely on its bag of visual words.

\subsection{Quantitative Comparison}

In this section, we describe our quantitative comparison between
SupDocNADE, DocNADE and sLDA. We used the implementation of sLDA
available at \url{http://www.cs.cmu.edu/~chongw/slda/} in our
comparison. For models which did not have a publicly available
implementation (hybrid topic/neural network model~\cite{wanhybrid} and
MMLDA~\cite{wang2011max}), we compare instead with the results reported in the
literature.

\subsubsection{Image Classification}
\label{sec:classification}

%

\begin{figure}
\minipage{\textwidth}
  \includegraphics[width=0.48\linewidth]{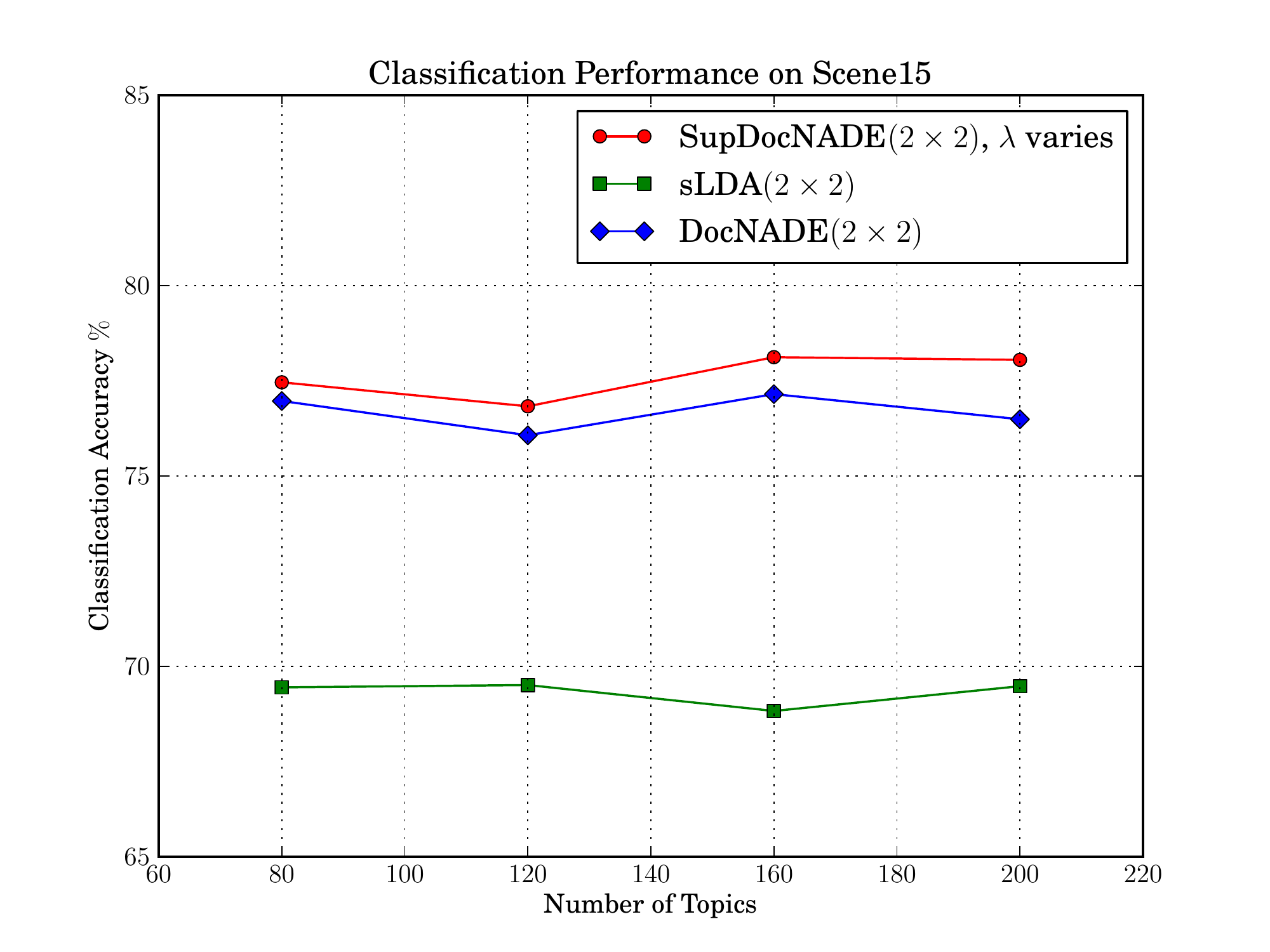}
  \includegraphics[width=0.48\linewidth]{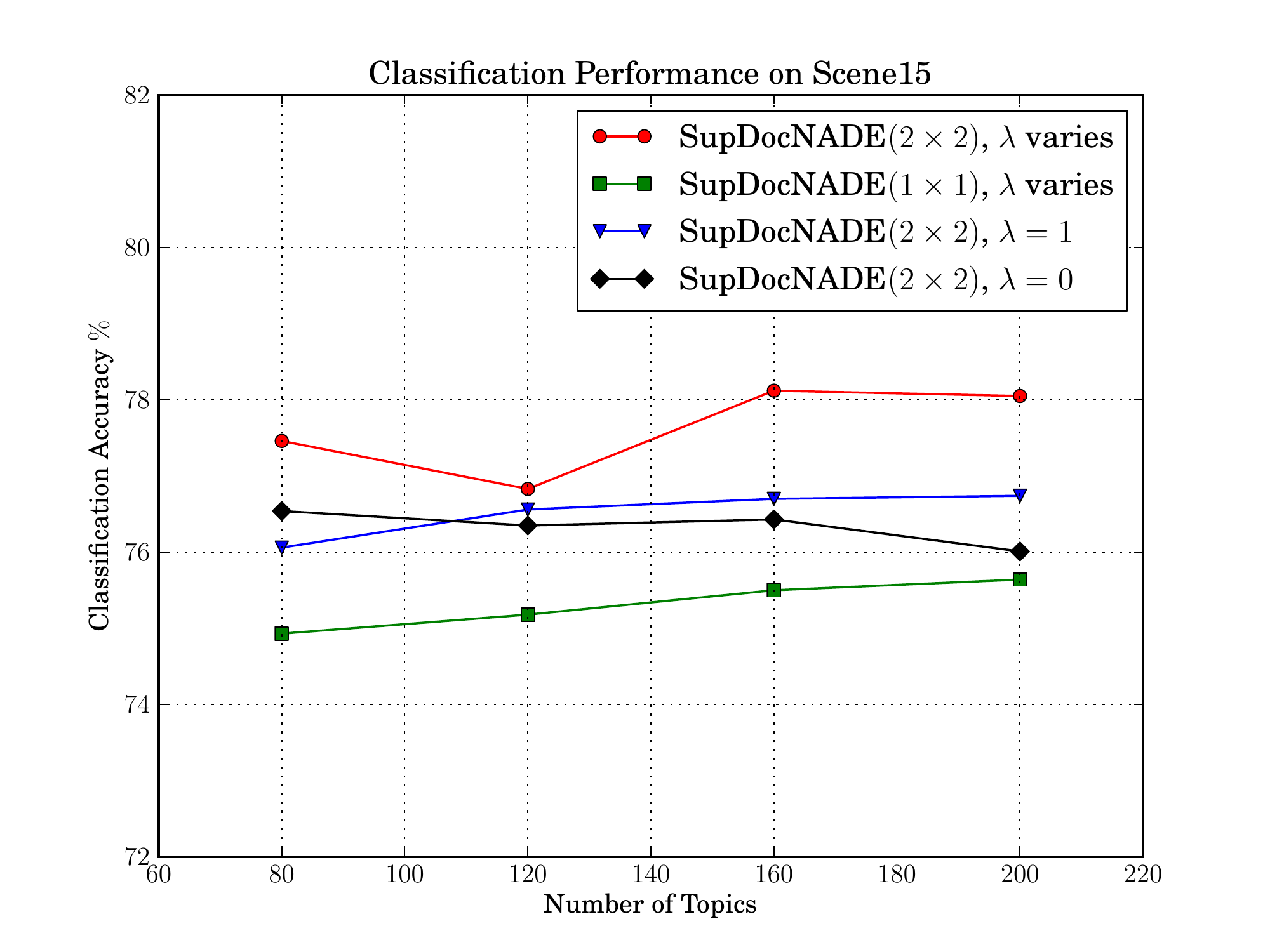}
\endminipage \hfill
\caption{Classification performance comparison on Scene15 dataset. The left figure shows the performance comparison between SupDocNADE, DocNADE and sLDA. The figure on the right compares the performance of different variants of SupDocNADE. }
\label{fig:scene_comp}
\end{figure}

\begin{figure}
\minipage{\textwidth}
  \includegraphics[width=0.48\linewidth]{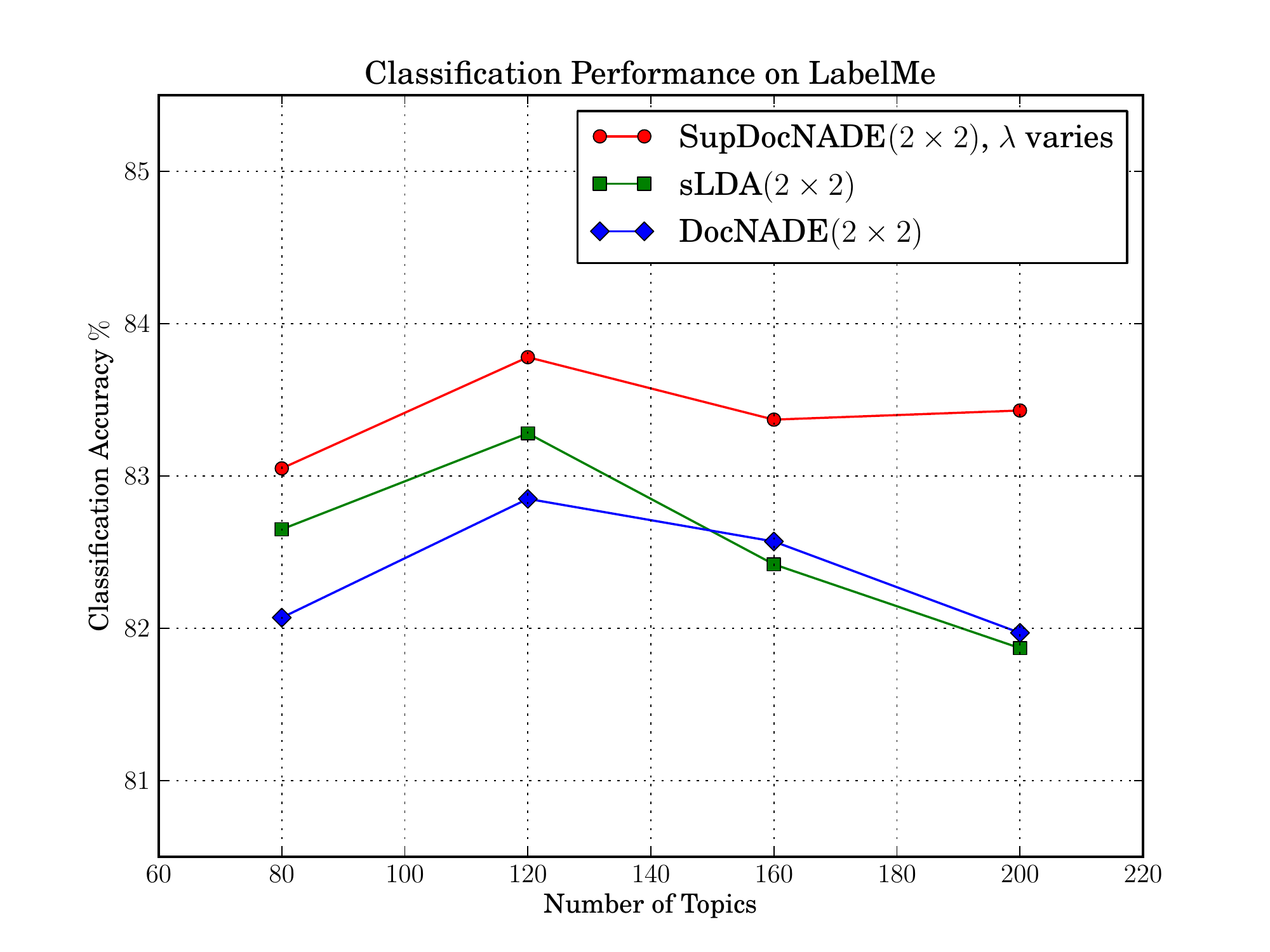}
  \includegraphics[width=0.48\linewidth]{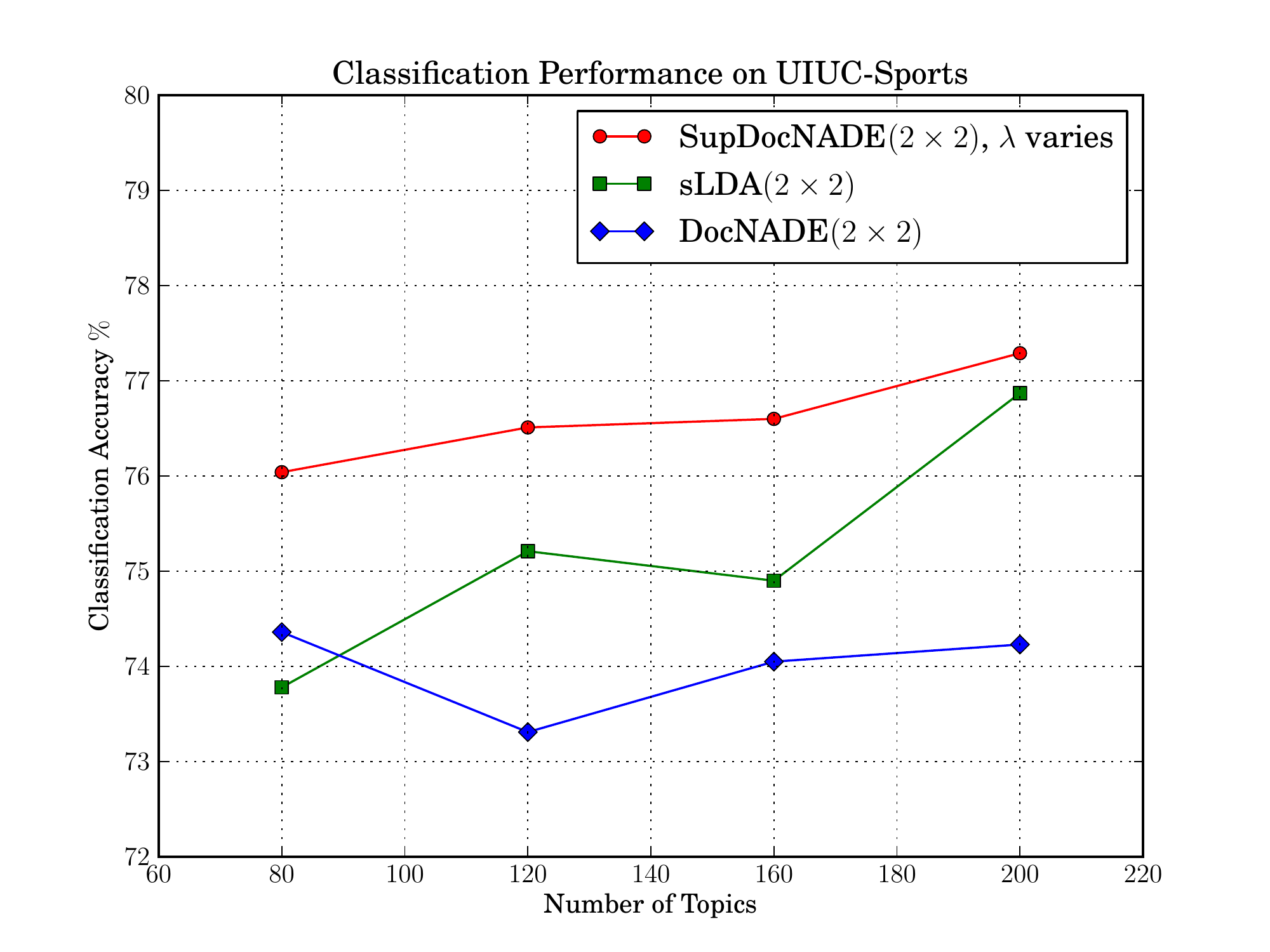}\\
  \includegraphics[width=0.48\linewidth]{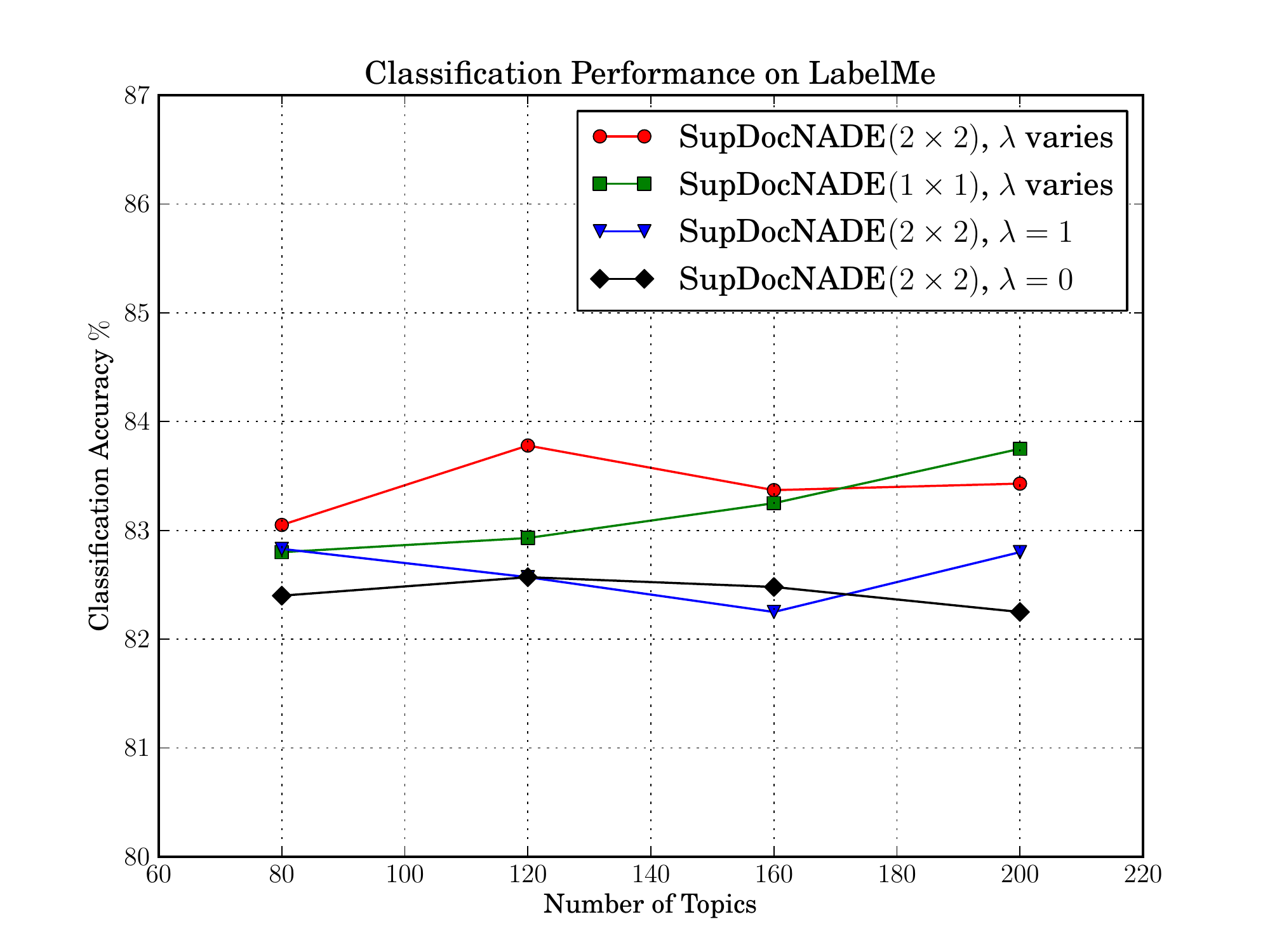}
  \includegraphics[width=0.48\linewidth]{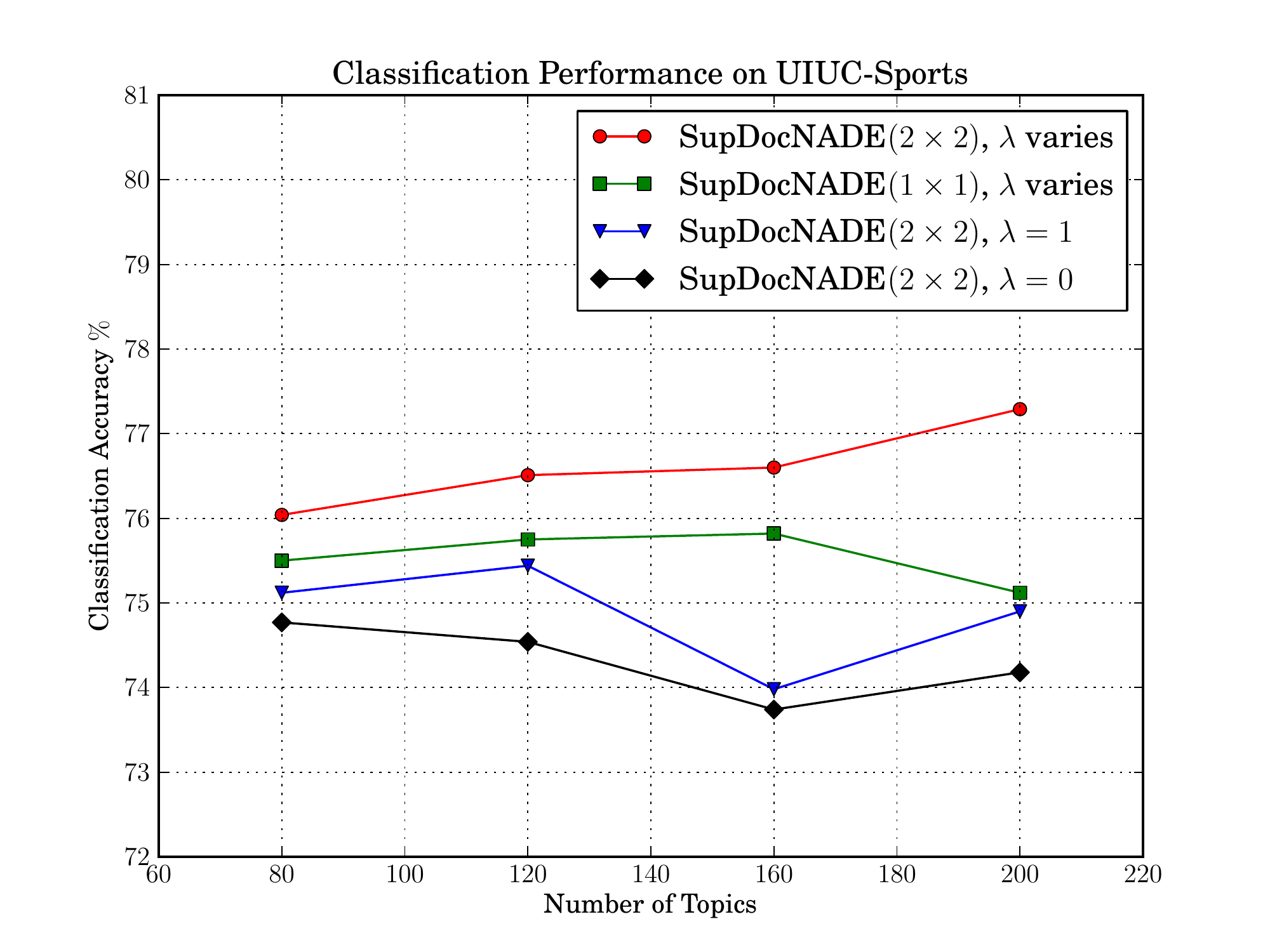}
\endminipage \hfill
\caption{Classification performance comparison on LabelMe and UIUC-Sports datasets. In the top row, we compare the classification performance of SupDocNADE, DocNADE and sLDA. In the bottom row, we compare the performance between different variants of SupDocNADE. Results on LabelMe are on the left and results on UIUC-Sports are on the right. }
\label{fig:labelme_and_uiuc_comp}
\end{figure}

We first test the classification performance of our method on the
Scene15 dataset. Figure~\ref{fig:scene_comp} illustrates the
performance of all methods, for a varying number of topics.
We observe that SupDocNADE
outperforms sLDA by a large margin, while also improving
over the orignal DocNADE model.

In Figure~\ref{fig:scene_comp}, we also compare
SupDocNADE with other design choices for the model,
such as performing purely generative ($\lambda=1$)
or purely discriminative ($\lambda=0$) training, or
ignoring spatial position information (i.e.\ using a single region,
covering the whole image). We see that both using position
information and tuning the weight $\lambda$ are important,
with pure discriminative learning performing worse.

\citet{wanhybrid} also performed experiments on the Scene15 dataset using
their hybrid topic/neural network model, but used a slightly
different setup: they used 45 topics, a visual word vocabulary of size 200,
a dense SIFT patch size of $32 \times 32$ and a step size of 16. They
also didn't incorporate spatial position information using a spatial grid.
When running SupDocNADE using this configuration, 
we obtain a classification accuracy of $73.36\%$, 
compared to $70.1\%$ for their model.


\subsubsection{Simultaneous Classification and Annotation}
\label{sec:classificatino and anno}

We now look at the simultaneous image classification and annotation
performance on LabelMe and UIUC-Sports datasets. 

The classification results are illustrated in
Figure~\ref{fig:labelme_and_uiuc_comp}. Similarly, we
observe that SupDocNADE outperforms DocNADE and sLDA. Tuning the
trade-off between generative and discriminative learning and
exploiting position information is usually beneficial. There
is just one exception, on LabelMe, with 200 hidden topic units, where
using a $1\times 1$ grid slightly outperforms a $2\times 2$ grid.

As for image annotation, we computed the performance
of our model with 200 topics. SupDocNADE obtains an
$F$\textup{-measure} of $43.87\%$  
and $46.95\%$ 
on the LabelMe and UIUC-Sports datasets respectively. This is
slightly superior to regular DocNADE, which obtains
$43.32\%$ and $46.38\%$.
Since code for performing image annotation using sLDA 
is not publicly available, we compare directly with the results found
in the corresponding paper~\cite{wang2009simultaneous}. \citet{wang2009simultaneous}
report $F$\textup{-measures} of $38.7\%$ and $35.0\%$ for sLDA,
which is below SupDocNADE by a large margin. 

We also compare with MMLDA~\cite{wang2011max}, a max-margin
formulation of LDA, that has been applied to image classification and
annotation separately. The reported classification accuracy for MMLDA
is $81.47\%$ (for LabelMe) and $74.65\%$ (for UIUC-Sports), which is
less than SupDocNADE. As for annotation,
$F$\textup{-measures} of $46.64\%$ (for LabelMe) and $44.51\%$ (for
UIUC-Sports) are reported, which is better than SupDocNADE on LabelMe
but worse on UIUC-Sports. We should mention that MMLDA did not address
the problem of {\it simultaneously} classifying and annotating images, these tasks being
treated separately.

\begin{figure}
\begin{center}
	\includegraphics[width=.93\linewidth]{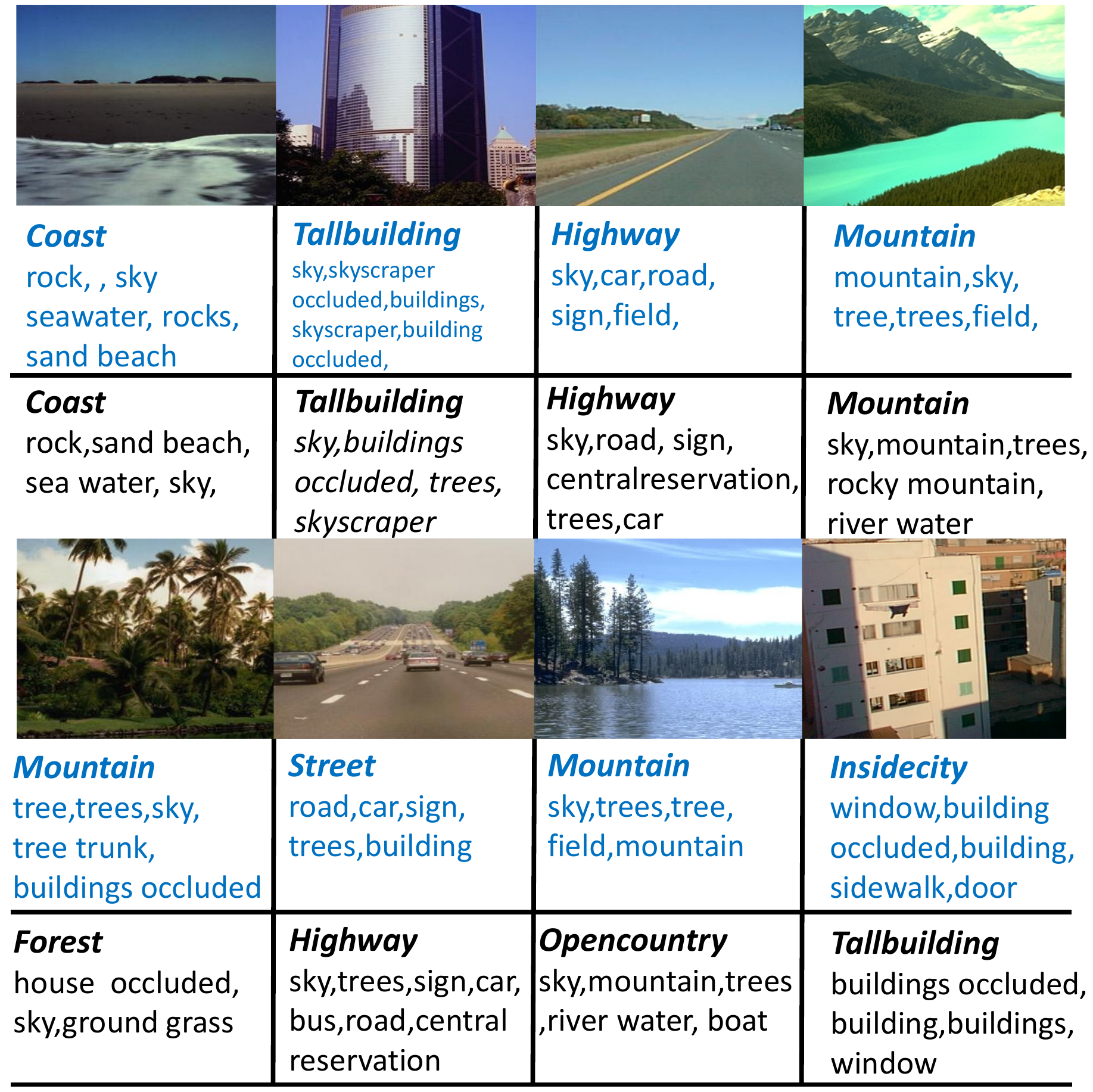}
\end{center}
\caption{Predicted class and annotation by SupDocNADE on LabelMe
  dataset. We list some correctly (top row) and incorrectly (bottom
  row) classified images. The predicted (in blue) and ground-truth (in
  black) class labels and annotation words are presented under each image.}
\label{fig:figure_results}
\end{figure}

Figure~\ref{fig:figure_results} illustrates examples of correct and incorrect
predictions made by SupDocNADE on the LabelMe dataset.

\subsection{Visualization of Learned Representations}
\label{sec:visualization of representation}
Since topic models are often used to interpret and explore the semantic structure
of image data, we looked at how we could observe the structure learned
by SupDocNADE. 

\begin{figure}[t]
\begin{center}
	\includegraphics[width=1\linewidth]{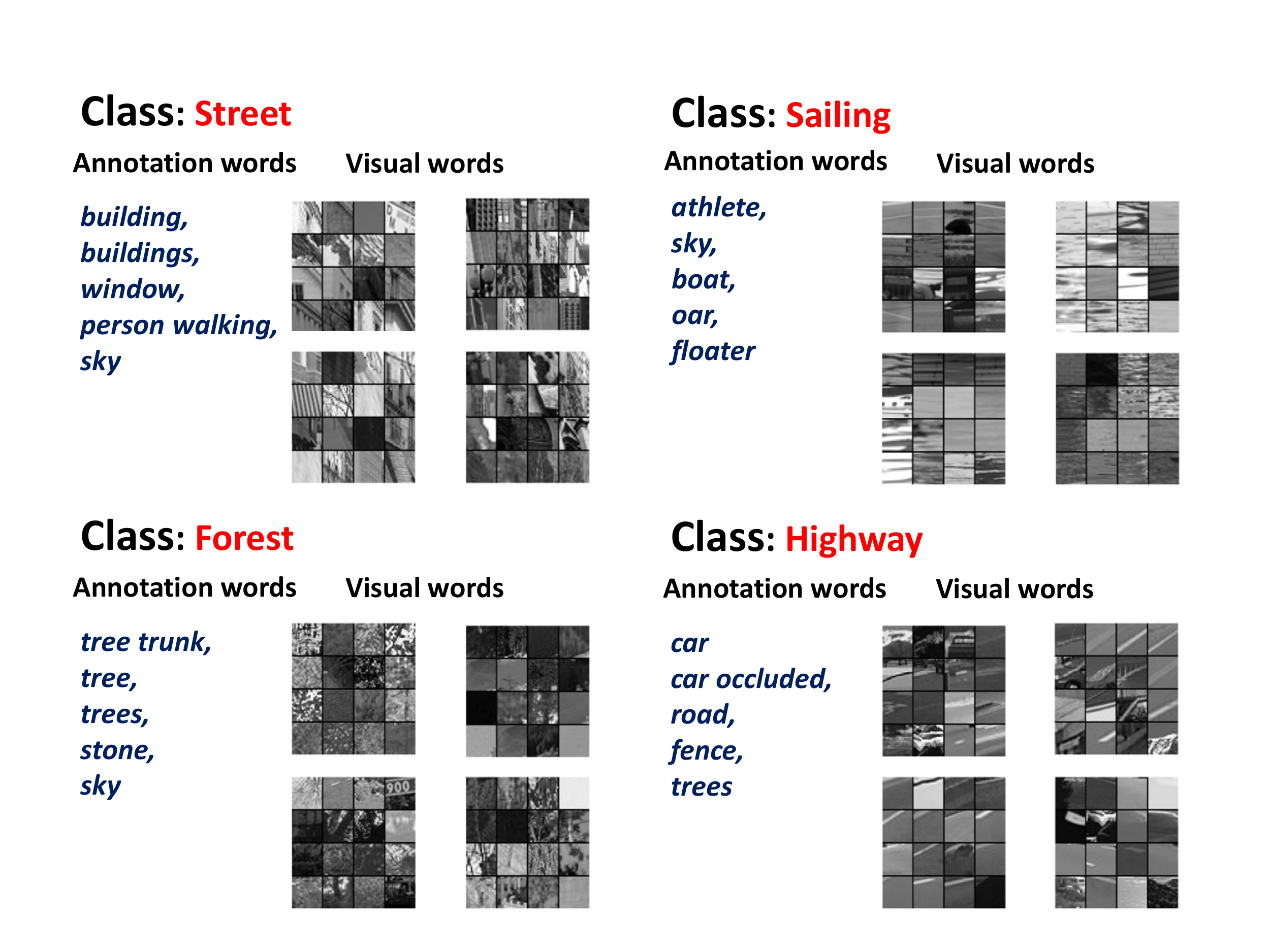}
\end{center}
\caption{Visualization of learned representations. Class labels are
  colored in red. For each class, we list 4 visual words (each represented by
  16 image patches) and 5
  annotation words that are strongly associated with each class. See
  Section~\ref{sec:visualization of representation} for more
  details. }
\label{fig:label_visual_anno}
\end{figure}

We tried to extract the visual/annotation words that were
most strongly associated with certain class labels within
SupDocNADE. For example, given a class label {\it street}, which
corresponds to a column $\mathbf{U}_{:,i}$ in matrix $\mathbf{U}$, we
selected the top 3 topics (hidden units) having the largest connection
weight in $\mathbf{U}_{:,i}$. Then, we averaged the columns of matrix
$\mathbf{W}$ corresponding to these 3 hidden topics and selected the
visual/annotation words with largest averaged weight connection.
The results of this procedure for classes {\it street}, {\it sailing},
{\it forest} and {\it highway} is illustrated in Figure~\ref{fig:label_visual_anno}.
To visualize the visual words, we show 16
image patches belonging to each visual word's cluster, as extracted
by $K$-means. 
The learned associations are intuitive: for example, the
class {\it street} is associated with the annotation words ``{\it building}'',
``{\it buildings}'', ``{\it window}'', ``{\it person walking}'' and ``{\it sky}'', while the visual words
showcase parts of buildings and windows.


%

%
\section{Conclusion and Discussion}
\label{conclusion}

In this paper, we proposed SupDocNADE, a supervised extension of
DocNADE. Like all topic models, our model is trained to model the
distribution of the bag of words representation of images and can
extract a meaningful representation from it. Unlike most topic models
however, the image representation is not modeled as a latent random
variable in a model, but instead as the hidden layer of a neural
network. While the resulting model might be less interpretable (as
typical with neural networks), it has the advantage of not requiring
any iterative, approximate inference procedure to compute an image's
representation.  Our experiments confirm that SupDocNADE is a
competitive approach for the classification and annotation of images.

\bibliographystyle{IEEEtranNAT}
\bibliography{egbib}

\end{document}